\documentclass[10pt,twocolumn,letterpaper]{article}

\usepackage[pagenumbers]{iccv} 

\input{preamble}

\definecolor{iccvblue}{rgb}{0.21,0.49,0.74}
\usepackage[pagebackref,breaklinks,colorlinks,allcolors=iccvblue]{hyperref}

\def\paperID{***} 
\def\confName{ICCV}
\def\confYear{2025}

\title{QArtSR: Quantization via Reverse-Module and Timestep-Retraining \\ in One-Step Diffusion based Image Super-Resolution}

\author{
\textbf{Libo Zhu}\textsuperscript{1}\thanks{Equal contribution.},
\textbf{Haotong Qin}\textsuperscript{2}\footnotemark[1], 
\textbf{Kaicheng Yang}\textsuperscript{1}, 
\textbf{Wenbo Li}\textsuperscript{4}, \\
\textbf{Yong Guo}\textsuperscript{5},
\textbf{Yulun Zhang}\textsuperscript{1}\thanks{Corresponding author: Yulun Zhang},
\textbf{Susanto Rahardja}\textsuperscript{3},
\textbf{Xiaokang Yang}\textsuperscript{1} \\
\textsuperscript{1}Shanghai Jiao Tong University, \textsuperscript{2}ETH Zürich,
\textsuperscript{3}Singapore Institute of Technology,
\\
\textsuperscript{4}Chinese University of Hong Kong,
\textsuperscript{5}Max Planck Institute for Informatics
}

\def\abstract{%
   \iftoggle{iccvpagenumbers}{}{%
     \thispagestyle{empty}
   }
   \centerline{\large\bf Abstract}%
   \vspace*{0pt}\noindent%
   \it\ignorespaces%
}

\begin{document}
\maketitle

\begin{abstract}
One-step diffusion-based image super-resolution (OSDSR) models are showing increasingly superior performance nowadays. However, although their denoising steps are reduced to one and they can be quantized to 8-bit to reduce the costs further, there is still significant potential for OSDSR to quantize to lower bits. To explore more possibilities of quantized OSDSR, we propose an efficient method, \textbf{Q}uantization vi\textbf{A} \textbf{r}everse-module and \textbf{t}imestep-retraining for OSD\textbf{SR}, named \textbf{QArtSR}. Firstly, we investigate the influence of timestep value on the performance of quantized models. Then, we propose Timestep Retraining Quantization (TRQ) and Reversed Per-module Quantization (RPQ) strategies to calibrate the quantized model. Meanwhile, we adopt the module and image losses to update all quantized modules. We only update the parameters in quantization finetuning components, excluding the original weights. To ensure that all modules are fully finetuned, we add extended end-to-end training after per-module stage. Our 4-bit and 2-bit quantization experimental results indicate that QArtSR obtains superior effects against the recent leading comparison methods. The performance of 4-bit QArtSR is close to the full-precision one. Our code will be released at \url{https://github.com/libozhu03/QArtSR}.
\end{abstract}

\setlength{\abovedisplayskip}{2pt}
\setlength{\belowdisplayskip}{2pt}

\vspace{-7.5mm}
\section{Introduction}
\vspace{-2.5mm}
Image super-resolution (SR) aims to reconstruct high-resolution (HR) images from low-resolution (LR) inputs by recovering fine details. Recent diffusion-based SR methods~\cite{stablesr,diffbir,seesr} can capture intricate data distributions and produce HR images with exceptional perceptual quality by modeling high-dimensional structures more effectively. They perform better than numerous earlier traditional methods~\cite{ISR_using_conv,Zhang_2018_CVPR,chen2022CAT,chen2023DAT}, which focus on simple synthetic degradations (\eg, Bicubic downsampling). And GAN-based approaches~\cite{BSRGAN,swinir,Real-ESRGAN} often suffer from training instability and performance fluctuations in real-world scenarios.

\begin{figure}[t]
\scriptsize
\centering
\vspace{-3.5mm}
\begin{tabular}{c c c c c c c c}

\hspace{-0.44cm}

\begin{adjustbox}{valign=t}
\begin{tabular}{ccccc}
\includegraphics[width=0.09\textwidth]{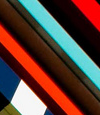} \hspace{-4mm} &
\includegraphics[width=0.09\textwidth]{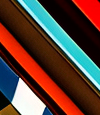} \hspace{-4mm} &
\includegraphics[width=0.09\textwidth]{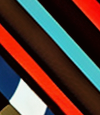} \hspace{-4mm} &
\includegraphics[width=0.09\textwidth]{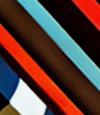} \hspace{-4mm} &
\includegraphics[width=0.09\textwidth]{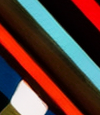}  \\

\includegraphics[width=0.09\textwidth]{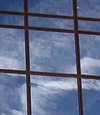} \hspace{-4mm} &
\includegraphics[width=0.09\textwidth]{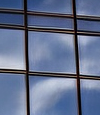} \hspace{-4mm} &
\includegraphics[width=0.09\textwidth]{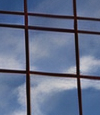} \hspace{-4mm} &
\includegraphics[width=0.09\textwidth]{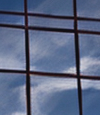} \hspace{-4mm} &
\includegraphics[width=0.09\textwidth]{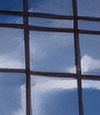}  \\

HR ($\times$4) \hspace{-4mm} &
DiffBIR~\cite{diffbir} \hspace{-4mm} &
OSEDiff~\cite{wu2024one} \hspace{-4mm} &
PassionSR~\cite{zhu2024passionsr} \hspace{-4mm} &
QArtSR \\
\# Step / Bits \hspace{-4mm} &
50 / 32-bit \hspace{-4mm} &
1 / 32-bit \hspace{-4mm} &
1 / 8-bit \hspace{-4mm} &
1 / 4-bit \\
Param. (M) \hspace{-4mm} &
1,618 \hspace{-4mm} &
1,303\hspace{-4mm} &
238\hspace{-4mm} &
122 \\
Ops (G) \hspace{-4mm} &
49,056 \hspace{-4mm} &
4,523 \hspace{-4mm} &
1,060 \hspace{-4mm} &
 531 \\
\end{tabular}
\end{adjustbox}
\end{tabular}
\vspace{-3.5mm}
\caption{Visual results of full-precision (FP) and low-bit multi-step and/or one-step diffusion SR models. Ops are computed with output size 512$\times$512. Compared to FP OSEDiff, QArtSR achieves about 90.66\% params reduction and 8$\times$ speedup.}
\label{fig:first-visual}
\vspace{-8mm}
\end{figure}

While diffusion-based models achieve high-quality results, they come with high latency, computational costs, and storage demands, limiting the deployment of hardware devices. Researchers have explored various lightweight strategies, particularly reducing denoising steps. Fast samplers~\cite{song2022denoisingdiffusionimplicitmodels, zhao2023unipcunifiedpredictorcorrectorframework} and distillation~\cite{song2023consistencymodels} significantly lower these steps. With advance of score distillation~\cite{yin2024onestepdiffusiondistributionmatching, wang2023prolificdreamerhighfidelitydiversetextto3d}, one-step diffusion SR (OSDSR) models, like SinSR~\cite{sinsr}, OSEDiff~\cite{wu2024one}, and DFOSD~\cite{DFOSD}, have become feasible. 

\begin{figure}[t]
\centering
\includegraphics[width=\columnwidth]{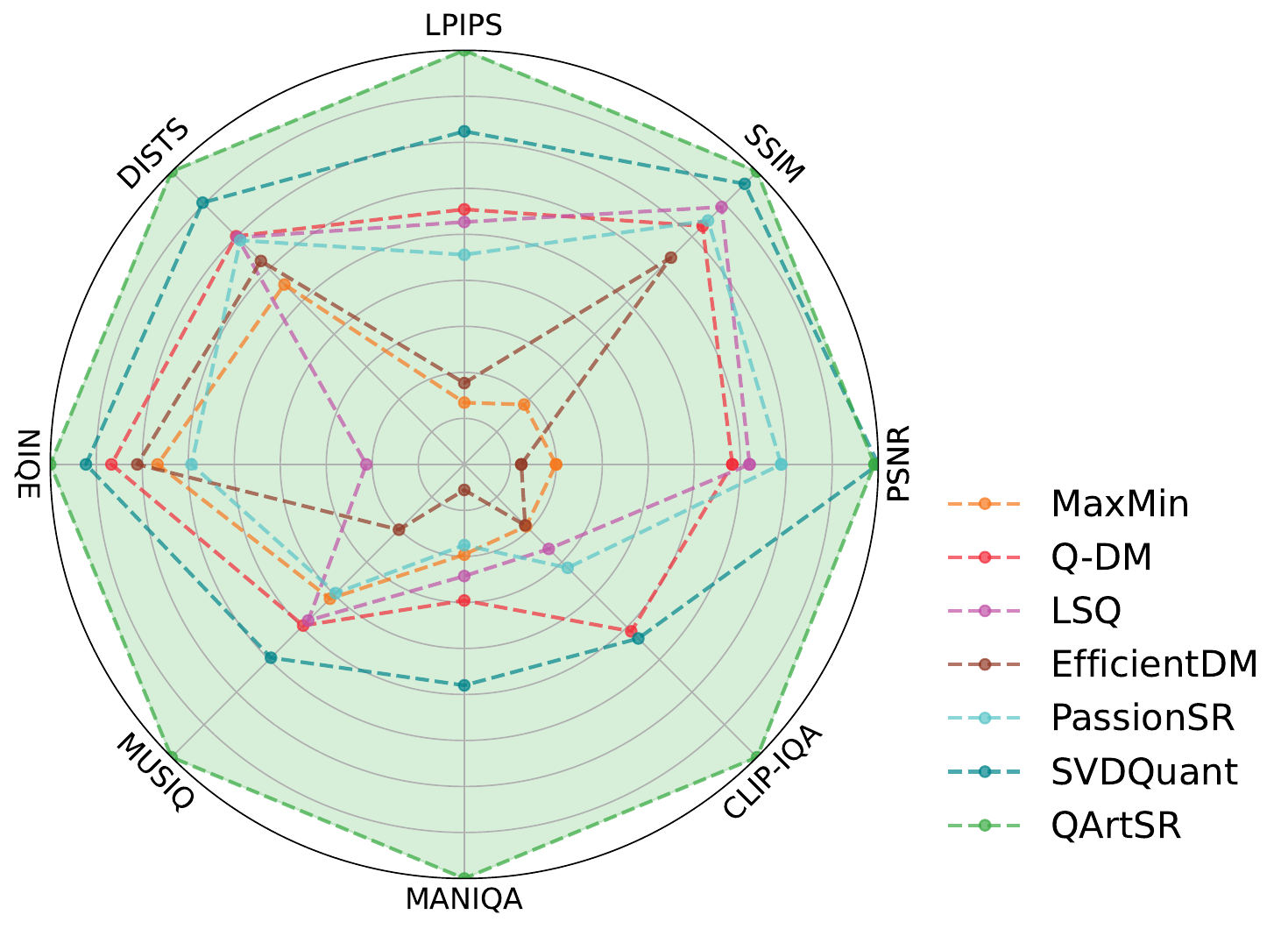}
\vspace{-8mm}
\caption{Performance visualization of low-bit quantization OSDSR methods at W4A4 bits setting on Urban100~\cite{Huang-CVPR-2015}.}
\label{fig:radar}
\vspace{-8mm}
\end{figure}

To further compress OSDSR models, model quantization~\cite{choukroun2019lowbitquantizationneuralnetworks, Ding_2022, pmlr-v139-hubara21a} stands out as an effective approach. Reducing activations and weights from full-precision (FP) to low-bit precision significantly decreases computational and storage costs. It is valuable under resource-constrained conditions. However, the inherent performance gap remains, necessitating strategies to minimize it for effective application in OSDSR models. PassionSR~\cite{zhu2024passionsr} is the first to introduce quantization to OSDSR, achieving 8-bit and 6-bit quantization. While 8-bit version maintains performance close to its FP counterpart (see Fig.~\ref{fig:first-visual}), the reconstruction quality degrades sharply at 4-bit or lower precision (see Fig.~\ref{fig:radar}). The ultra-low-bit quantization on both weights and activations has remained a challenging issue over time. It is hard for the ultra-low-bit quantized model to retain the performance.

Aiming at compressing the OSDSR further, we concentrate on addressing model quantization for ultra-low-bit settings. For multi-step diffusion's ultra-low-bit quantization, current leading quantization strategies~\cite{li2024q,He2023EfficientDM,li2024svdqunat} have achieved promising results at 4-bit. However, applying these methods to OSDSR models still leads to substantial performance degradation (see Fig.~\ref{fig:radar}). To improve model performance, we must address the following challenges:

\emph{\textbf{\Rmnum{1}. Extremely constrained expression capability in low-bit settings.}}  
Under ultra-low-bit settings, quantized models exhibit significantly weaker expression capability than the full-precision one. The higher low-bit settings (\eg, 8-bit) allow performance recovery through distillation with the full-precision counterpart. However, ultra-low-bit settings (\eg, 2$\sim$4-bit) struggle to generate high-quality images or even complete image generation tasks. Due to the inherent performance ceiling of quantized models, distillation or similar methods bring limited improvement. 

\emph{\textbf{\Rmnum{2}. Severe optimization difficulties in discrete space.}} 
Converting a full-precision model to its quantized counterpart involves transitions from a continuous to a discrete space. In ultra-low-bit settings, the initial state in the discrete space is significantly distant from the optimal solution. Moreover, inaccurate gradients and quantization errors further hinder convergence, making it challenging for the quantized model to narrow its performance drop. As a result, achieving satisfactory performance demands extensive training time and computational resources.

\emph{\textbf{\Rmnum{3}. Degradation of quantization for one-step features.}} According to our quantization experiments,  many quantization methods present outstanding performance on multi-step diffusion models while weak performance on one-step diffusion models. This means that the quantization of the one-step diffusion model is quite different from the previous diffusion quantization work. We need to fully analyze the theoretical reason behind this phenomenon to design a more suitable quantization method for OSDSR.

\begin{figure}[t]
\scriptsize
\centering
\vspace{-3.5mm}
\begin{tabular}{c c c c c c c c}

\hspace{-0.44cm}

\begin{adjustbox}{valign=t}
\begin{tabular}{ccccc}
\includegraphics[width=0.09\textwidth]{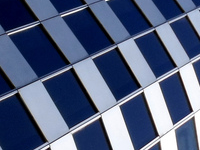} \hspace{-4mm} &
\includegraphics[width=0.09\textwidth]{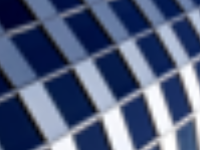} \hspace{-4mm} &
\includegraphics[width=0.09\textwidth]{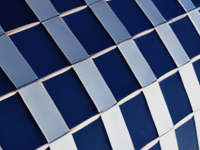} \hspace{-4mm} &
\includegraphics[width=0.09\textwidth]{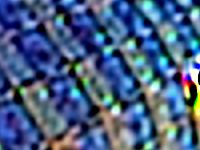}  \hspace{-4mm} &
\includegraphics[width=0.09\textwidth]{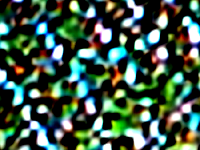}  \\
HR \hspace{-4mm} &
LR ($\times$4) \hspace{-4mm} &
OSEDiff~\cite{wu2024one} \hspace{-4mm} &
MaxMin~\cite{jacob2017quantizationtrainingneuralnetworks} \hspace{-4mm} &
LSQ~\cite{Bhalgat2020LSQ+} \\
\includegraphics[width=0.09\textwidth]{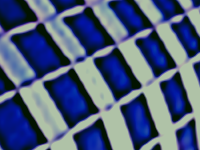} \hspace{-4mm} &
\includegraphics[width=0.09\textwidth]{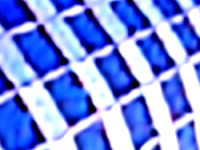} \hspace{-4mm} &
\includegraphics[width=0.09\textwidth]{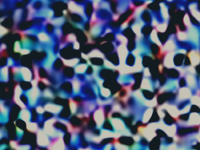} \hspace{-4mm} &
\includegraphics[width=0.09\textwidth]{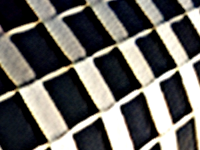} \hspace{-4mm} &
\includegraphics[width=0.09\textwidth]{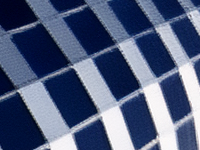}  \\
Q-DM~\cite{li2024q} \hspace{-4mm} &
EfficientDM~\cite{He2023EfficientDM} \hspace{-4mm} &
PassionSR~\cite{zhu2024passionsr} \hspace{-4mm} &
SVDQuant~\cite{li2024svdqunat} \hspace{-4mm} &
QArtSR (ours) \\
\end{tabular}
\end{adjustbox}
\end{tabular}
\vspace{-3.5mm}
\caption{Visual comparison ($\times$4) of 32-bit OSEDiff~\cite{wu2024one} and 2-bit quantized models with various quantization methods.}
\label{fig:second-visual}
\vspace{-6.5mm}
\end{figure}

To alleviate those challenges, we propose an effective and efficient quantization method,  \textbf{q}uantization vi\textbf{a} \textbf{r}everse-module and \textbf{t}imestep-retraining for one-step diffusion-based image \textbf{s}uper-\textbf{r}esolution, named QArtSR. 

We first explore the impact of timestep value on OSDSR quantization and propose the timestep retraining quantization (TRQ). It provides a better initial state for the quantized model, reducing the pressure of following the quantization process. Additionally, we propose a novel reversed per-module quantization (RPQ) strategy, whose quantization order is reversed to the inference sequence. By gradually introducing quantized modules, performance degradation can be recovered more effectively. The reversed order makes joint optimization of the image and module losses possible. We further apply extended training after the per-module stage, ensuring all modules are fully finetuned.

In the quantization process of our QArtSR, only the parameters in the specially designed finetuning components are updated with the combined loss, excluding the original weights. It is as efficient as recent post-training quantization methods, consuming similar GPU memory and time.

OSEDiff~\cite{wu2024one} is selected as our quantization foundation owing to its superior performance and high inference speed in QArtSR. Compared to 32-bit full-precision version OSEDiff~\cite{wu2024one}, QArtSR achieves an impressive \textbf{90$\sim$95\%} reduction in parameters and operations with 4$\sim$2 bits. In Figs.~\ref{fig:first-visual} and~\ref{fig:second-visual}, QArtSR maintains perceptual quality comparable to full-precision models at 4-bit and preserves most performance even at 2-bit. Figure~\ref{fig:radar} demonstrates its superiority over state-of-the-art diffusion quantization methods. Overall, our key contributions are as follows:
\begin{itemize}
    \item We propose an efficient ultra-low-bit quantization method for OSDSR, QArtSR. To the best of our knowledge, we are the first to quantize the OSDSR model to 2$\sim$4 bits on both the activation and weight efficiently.
    \item We propose a timestep retraining quantization (TRQ). We retrain the backbone with the best value of timestep, reducing the quantization error significantly. 
    \item We propose a reversed per-module quantization (RPQ), considering the partial and overall losses meanwhile. The quantization order is reversed to the inference sequence.
    \item Our QArtSR delivers perceptual quality nearly matching that of a full-precision model at 4-bit. It significantly outperforms other leading quantization methods in both performance and scores across 2$\sim$4-bit settings.
\end{itemize}

\section{Related Work}  
\vspace{-1mm}  
\subsection{Image Super-Resolution}  
\vspace{-1mm}  
Reconstructing high-resolution (HR) images from low-resolution (LR) inputs degraded by complex factors has long been a challenging task, forming the core objective of image super-resolution (SR). Early SR techniques~\cite{AISR, zhang2018imagesuperresolutionusingdeep, chen2023crossaggregationtransformerimage} and GAN-based approaches~\cite{BSRGAN, Real-ESRGAN, swinir} have significantly advanced the field. Recently, stable diffusion (SD) models~\cite{rombach2022highresolutionimagesynthesislatent} have emerged as powerful tools, leveraging strong generative priors and the ability to model the intricate data distributions. Notable works, such as DiffBIR~\cite{diffbir}, SeeSR~\cite{seesr}, and InvSR~\cite{yue2024arbitrary}, have further improved perceptual quality. However, due to their multi-step nature, substantial inference latency is introduced, limiting real-world applications. To overcome this, one-step diffusion SR (OSDSR) models (\eg, SinSR~\cite{sinsr}, OSEDiff~\cite{wu2024one}) have been proposed, significantly reducing the inference time by condensing the process into a single step.
\vspace{-1mm}
\subsection{Model Quantization}
\vspace{-1mm}
Model quantization enhances performance by reducing parameter precision while preserving effectiveness. Depending on whether weight retraining is involved, quantization methods are categorized into post-training quantization (PTQ)~\cite{ZeroQuant,brecq} and quantization-aware training (QAT)~\cite{Bhalgat2020LSQ+,efficientqat, yu2024improving}. However, the distinction between PTQ and QAT has become increasingly blurred with the introduction of minor finetuning in PTQ. Quantization has also been demonstrated as an effective compression technique for deploying large language models~\cite{liu2023llm, bondarenko2024low, li2023loftq, smoothquant, OmniQuant} on terminal devices.
\vspace{-1mm}
\subsection{Quantization of Diffusion Models}
\vspace{-1mm}
With the rapid advances of diffusion models (DM), researchers have increasingly focused on improving their efficiency through quantization. PTQ4DM~\cite{PTQ4DM} pioneers the study of quantized diffusion models, identifying key challenges. Subsequent PTQ methods, such as Q-Diffusion~\cite{q-diffusion} and PTQD~\cite{PTQD}, introduce specialized calibration strategies for diffusion models. Additionally, Q-DM~\cite{li2024q} presents the first QAT-based method for low-bit multi-step DM, further advancing quantized diffusion.
BitsFusion~\cite{sui2024bitsfusion} and BitDistiller~\cite{du2024bitdistiller} have made significant progress in exploring weight-only quantization. The feedforward layer is recognized as particularly sensitive to quantization and model performance can be improved by selectively retraining it in QuEST~\cite{QuEST}. Recently, EfficientDM~\cite{He2023EfficientDM} designs a low-rank quantization finetuning strategy and SVDQuant~\cite{li2024svdqunat} utilizes 16-bit parallel low-rank skip-connection to retain the performance for multi-step DMs. As for one-step diffusion-based SR (OSDSR) models, PassionSR~\cite{zhu2024passionsr} proposes a novel quantization strategy and first quantizes OSDSR to 8-bit and 6-bit. However, limited research has focused on the ultra-low-bit (\ie, 2$\sim$4-bit) quantization of OSDSR, which differs considerably from multi-step DM.


\section{Methods}
\vspace{-1.5mm}
\subsection{Preliminaries}
\vspace{-1.5mm}
\noindent \textbf{Diffusion Models.} Diffusion models~\cite{rombach2022highresolutionimagesynthesislatent} iteratively transform a complex data distribution into a simpler one by gradually adding noise. The reversed process, which reconstructs the data, is learned to generate new samples. The true data distribution is denoted as $ p_{\text{data}}(\mathbf{x})$. The forward diffusion process progressively turns the data $ \mathbf{x}_0 $ into pure noise $ \mathbf{x}_T $ through a series of steps, which is defined as:
\begin{equation}
q(\mathbf{x}_t | \mathbf{x}_{t-1}) = \mathcal{N}(\mathbf{x}_t; \sqrt{1 - \beta_t} \mathbf{x}_{t-1}, \beta_t \mathbf{I}),
\label{equ: diffusion_base}
\end{equation}
where $ \beta_t $ is a scheduling parameter controlling the noise level at each step, and $ t $ denotes the time step.

After $ T $ steps, the distribution converges to a standard normal form as: $ \mathbf{x}_T \sim \mathcal{N}(0, \mathbf{I}) $. The reversed diffusion process reconstructs the original data distribution by learning a denoising model $ p_{\theta}(\mathbf{x}_{t-1} | \mathbf{x}_t) $ as follows:
\begin{equation}
p_{\theta}(\mathbf{x}_{t-1} | \mathbf{x}_t) = \mathcal{N}(\mathbf{x}_{t-1}; \mu_{\theta}(\mathbf{x}_t, t), \sigma^2(t) \mathbf{I}),
\label{equ: diffusion_base2}
\end{equation}
where $ \mu_{\theta} $ represents the denoising mean learned by a neural network to approximate the original data distribution, and $ \sigma^2(t) $ denotes the noise variance at step $ t $. By repeatedly applying this denoising step, the model progressively refines the noisy inputs, ultimately generating high-quality samples from an initial random noise distribution.

\noindent \textbf{Model Quantization.} Model quantization~\cite{jacob2017quantizationtrainingneuralnetworks} reduces memory and computational costs by converting model parameters and activations into lower-bit integers. This process utilizes a scale $ s $ and zero-point bias $ z $ to compensate for data distribution shifts. For a floating-point vector $ \mathbf{x} $, the quantization and dequantization operations are defined as:
\begin{equation}
\begin{cases} 
    x_{\text{int}} = \text{CLIP}\left(\frac{\mathbf{x}-z}{s}, l, u \right) \\
    \hat{x} = s \cdot x_{\text{int}} + z
\end{cases},
\label{equ: q_base}
\end{equation}
where $x_{\text{int}}$ refers to the quantized integer form and $\hat{x}$ is simulated quantized form. $ s $ controls quantization precision, and $ z $ adjusts the data offset before scaling. $ \text{CLIP}(\cdot, l, u) $ ensures values remain within the specified range $[l, u]$.

Due to the non-differentiability of rounding in quantization,  we utilize the straight-through estimator (STE~\cite{liu2022nonuniform}) to approximate gradients for the quantization training:
\begin{equation}
\frac{\partial L(x)}{\partial x} \approx 
\begin{cases} 
1 & \text{if } x \in [l, u] \\
0 & \text{otherwise}.
\end{cases},
\label{equ:STE}
\end{equation}

\noindent \textbf{Equivalent Transformation.} Outliers in weight and activation distribution have negative effects on quantization, decreasing the quantized model performance largely. For a linear layer with weight $W$, activation $X$ and bias $B$, we can conduct equivalent transformation with scale
$\phi$ and off-set $\gamma$ to change quantizers' target to $\hat{W}$ and $\hat{X}$ as~\cite{zhu2024passionsr}:
\begin{equation}
 \tilde{X} = (X - \delta) \oslash \phi, \tilde{W} = \phi \odot W, \tilde{B} = B + \gamma W,
\label{equ: linear transformation}
\end{equation}
where $\odot$, $\oslash$ are element-wise multiplication and division. ET adjusts weight and activation distribution and is an effective method to eliminate outliers for quantization~\cite{smoothquant}. 

\begin{figure*}[t]
\centering
\includegraphics[width=0.97\textwidth]{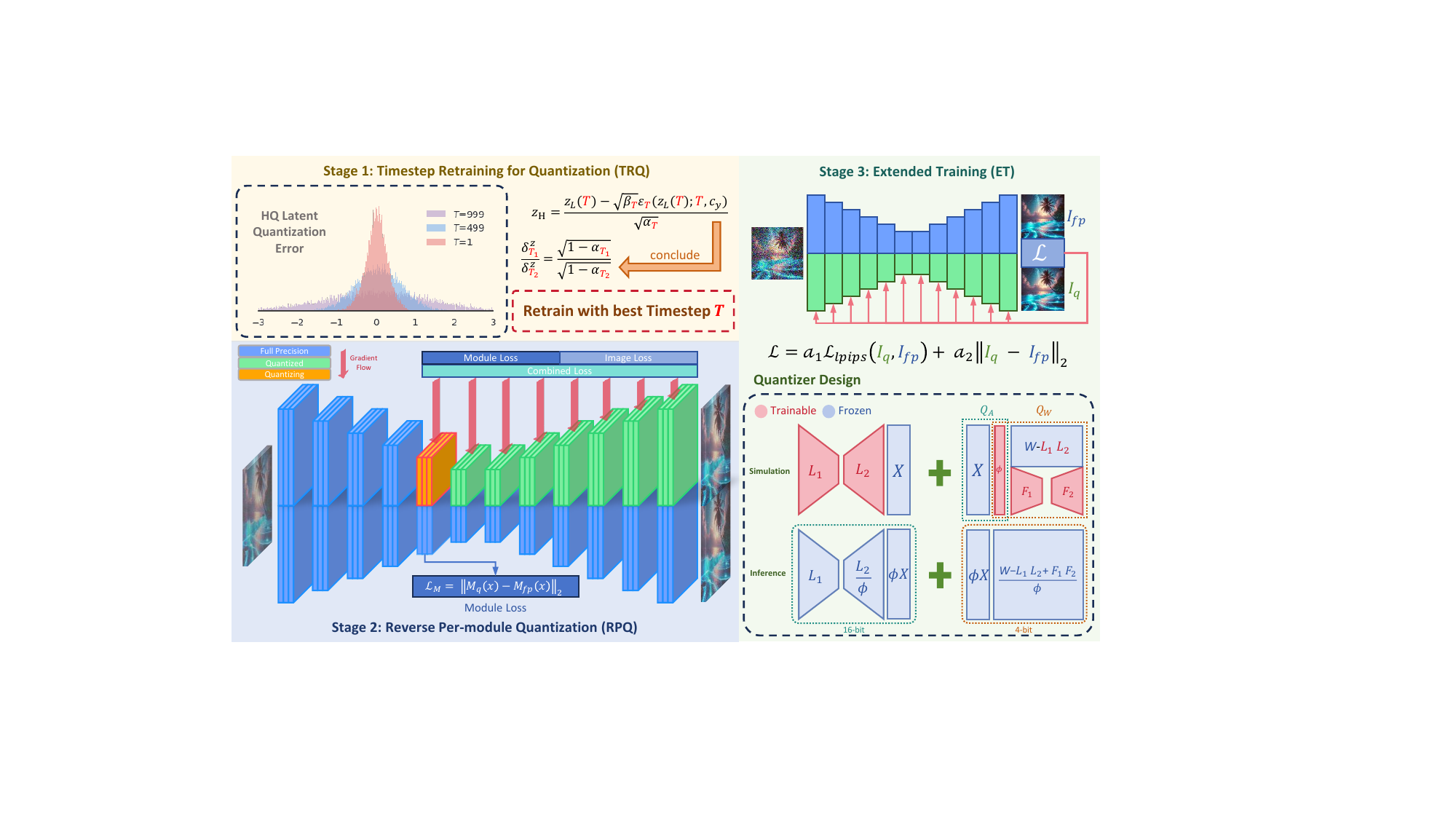}
\vspace{-3mm}
\caption{Overview of our QArtSR. Stage 1: we research the relationship between timestep and quantization error. We retrain the OSDSR with the best timestep $T$ before quantization. Stage 2: we propose a reversed per-module quantization strategy to make the process of quantization finetuning more smooth. Stage 3: we need to carry on the extended end-to-end training to enhance the performance further.}
\label{fig:overview}
\vspace{-6mm}
\end{figure*}

\vspace{-1mm}
\subsection{Timestep Retraining for Quantization (TRQ)}
\vspace{-2mm}
As mentioned in OSEDiff~\cite{wu2024one}, the LR-to-HR latent transformation is commonly used in one-step diffusion-based image super-resolution (OSDSR) models, connecting the UNet and VAE decoder. It has a great impact on model quantization due to the selection of timesteps.

The LR-to-HR latent transformation $F_{T}$ is formulated as a text-conditioned image-to-image denoising process in Eq.~\eqref{equ: OSDSR-LH}. Utilizing the predicting noise function $\epsilon_{T}$ of Unet, it conducts only one-step denoising on the LR latent $Z_L$ to obtain the HR latent $Z_H$ at the $T$-th diffusion timestep. 
\begin{equation}
{Z}_H = F_{T}(Z_L;c_y)=\frac{Z_L-\sqrt{\beta_T}\epsilon_{T}(Z_L;T,c_y)}{\sqrt{\alpha_T}},
\label{equ: OSDSR-LH}
\end{equation}
where ${\alpha}_T$ and ${\beta}_T$ are the scalars that are dependent to the timestep $T$ and they satisfy the condition, $\beta_T = 1 - \alpha_T$. To demonstrate the influence of timestep $T$ on the quantization error, we introduce the theorem below.

\begin{figure}[t]
\centering
\includegraphics[width=\columnwidth]{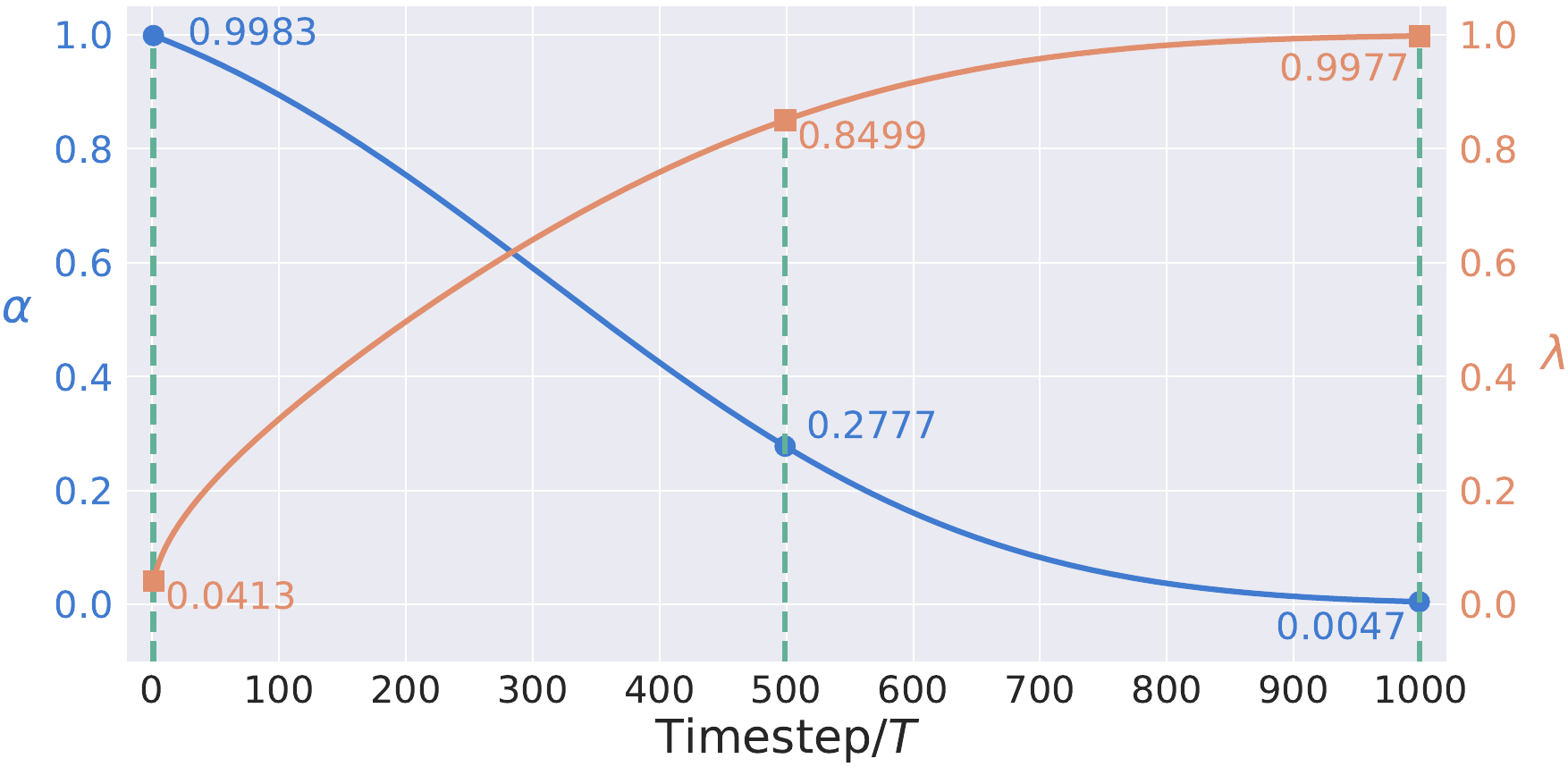}
\vspace{-7.5mm}
\caption{The value of $\alpha$ and $\lambda$ of different timestep $T$.}
\label{fig:t_study}
\vspace{-8mm}
\end{figure}

\noindent \textbf{Theorem: The high-resolution (HR) latent $Z_H$'s quantization error $\delta^Z(T)$ is proportional to $\lambda=\sqrt{1 - \alpha_T}$.} 
\begin{equation}
   \delta^Z(T) \propto \lambda=\sqrt{1 - \alpha_T}.
\label{equ:theorem}
\end{equation}
We provide the detailed proof for this theorem in the supplementary material. We display value of $\alpha$ and $\lambda$ as timestep $T$ varies in Fig.~\ref{fig:t_study}. Timestep $T$ selection plays a crucial role in reducing HR latent quantization error and improving the HR image quality. It is obvious that $T = 1,000$, which OSEDiff takes, leads to the largest quantization error while $T = 1$ leads to the smallest quantization error. 

To prove our theorem, we also provide the experimental results of different timesteps (1,000, 500, and 1) models under different bits settings (W32A32, W8A8, W4A4, and W2A2) in Tab.~\ref{tab:Step Study}. The model performances of different timesteps are evenly matched under the full-precision setting. The model of the smaller timestep tends to demonstrate stronger performance as the bits number decreases.

It is worth noting that the timestep $T$ has nothing to do with the number of denoising steps of diffusion-based models. We only concentrate on how the value of timestep $T$ impacts quantization while there is still only one denoising step in our full-precision backbone model.

In order to minimize the quantization error, we select $T = 1$ as the best timestep and retrain the model with the best timestep as our full-precision backbone. This retrained model with $T = 1$ provides a strong foundation for further quantization experiments and applications.

\begin{table*}[t]
\centering
\scriptsize
\resizebox{1\textwidth}{!}{
\begin{tabular}{c | c | c | c c c c c c c c}
\toprule
\rowcolor{iccvblue!30}
{Datasets} & {Bits} & {Timestep $T$}  & PSNR$\uparrow$  & SSIM$\uparrow$ & LPIPS$\downarrow$ & DISTS$\downarrow$ & NIQE$\downarrow$ & MUSIQ$\uparrow$ & MANIQA$\uparrow$ & CLIP-IQA$\uparrow$ \\
\midrule
&& 1,000 & 25.27	&0.7379	&0.3027	&\textcolor{red}{0.1808}	&4.355	&67.43	&\textcolor{red}{0.4766}	&0.6835\\
&& 500  & \textcolor{red}{26.11}	& \textcolor{red}{0.7442}	& \textcolor{red}{0.2898}	&0.1820	&4.414	&\textcolor{red}{67.73}	&0.4551	&\textcolor{red}{0.6939} \\
&\multirow{-3}{*}{W32A32} & 1 &25.39	&0.7195	&0.3251	&0.1918	&\textcolor{red}{4.246}	&67.59	&0.4649	&0.6859\\
\cline{2-11}
&& 1,000 &25.26  &0.7298	&0.3202	&0.1862	&4.334	&66.82	&\textcolor{red}{0.464}	&0.6854\\
&& 500 & \textcolor{red}{26.15}	&0.7444	&0.2929	&0.1835	&4.406	&\textcolor{red}{67.43}	&0.4526	&\textcolor{red}{0.6916}\\
&\multirow{-3}{*}{W8A8} & 1 & 26.03	&\textcolor{red}{0.7499}	&\textcolor{red}{0.2806}	&\textcolor{red}{0.1726}	&\textcolor{red}{4.234}	&66.93	&0.4458	&0.6707 \\
\cline{2-11}
&& 1,000 &16.01	&0.5033	&0.7016	&0.3668	&\textcolor{red}{5.520}	&31.26	&\textcolor{red}{0.2771}	&0.3099\\
&& 500 &24.16	&0.6415	&0.5963	&0.386	&6.855	&35.47	&0.2411	&0.3069\\
&\multirow{-3}{*}{W4A4} & 1 &\textcolor{red}{26.31}	&\textcolor{red}{0.7328}	&\textcolor{red}{0.4423}	&\textcolor{red}{0.3035}	&7.280	&\textcolor{red}{38.49}	&0.2010	&\textcolor{red}{0.3941}  \\
\cline{2-11}
&& 1,000 &14.88	&0.2181	&0.7263	&0.4584	&8.525	&33.12	&0.2187	&0.2190\\
&& 500 & 21.02	&0.6398	&0.4586	&0.2849	&8.292	&36.96	&\textcolor{red}{0.2382}	&0.2255\\
\multirow{-12}{*}{RealSR} & \multirow{-3}{*}{W2A2} & 1 & \textcolor{red}{26.01}	& \textcolor{red}{0.7453}	& \textcolor{red}{0.4462}	& \textcolor{red}{0.2741}	&\textcolor{red}{7.455}	&\textcolor{red}{37.46}	&0.1826	&\textcolor{red}{0.2932} \\
\cline{1-11}
\bottomrule
\end{tabular}
}
\vspace{-3mm}
\caption{MaxMin quantization experiments ($\times$4) of OSEDiff~\cite{wu2024one} on RealSR~\cite{Ji_2020_CVPR_Workshops} under different timestep $T$ and bits settings. W$w$A$a$ denotes $w$ bits weight and $a$ bits activation quantization. The best results in the same setting are colored with \textcolor{red}{red}.}
\label{tab:Step Study}
\vspace{-5.5mm}
\end{table*}

\vspace{-0.7mm}
\subsection{Reversed Per-module Quantization (RPQ)}
\vspace{-1mm}
In the ultra-low-bit settings, quantizing all modules leads to significant performance degradations, and end-to-end training struggles to recover the model's performance. Providing a better starting point for finetuning allows faster convergence with the reduced time and GPU resource consumption. Inspired by EfficientQAT~\cite{efficientqat}, we propose a reversed per-module quantization strategy by reversing quantization order compared to the forward inference sequence.

Original per-module quantization finetuning focuses on partial optimization, often neglecting overall performance. Even if each quantized module achieves minimal error, the overall image quality may not be optimal. Therefore, we also need to concentrate on the quality of the final quantized image while training certain quantized modules.

The image and module losses are designed as follows:
\begin{equation}
\left\{
\begin{aligned}
L_{\text{image}} &= a_1L_{\text{lpips}}(I_q, I_{fp}) + a_2||I_q - I_{fp}||_2 \\
L_M &= || M_q(x) - M_{fp}(x)||_2 
\end{aligned}
\right.,
\label{equ:loss_design}
\end{equation}
where $I_q$, $I_{fp}$ indicate the quantized, full-precision image and $M_q$, $M_{fp}$ indicate the quantized, full-precision module. $\Vert \cdot \Vert_2$ indicates the mean square error (MSE) loss. $L_{lpips}$ refers to the reference-based evaluation metrics LPIPS~\cite{zhang2018unreasonable}. $a_1$ and $a_2$ are weighting factors of the two losses.

To consider module and image losses simultaneously, we reverse the forward inference module sequence as the per-module quantization order. Starting from the module closest to output obtains more faithful quantized images, which reflects the current quantized model's performance better. If the quantization order is the same as the inference sequence, there are plenty of FP modules behind the quantized modules. They may recover the performance of quantized modules to some extent, so that the quantized image may not align with the performance of the model.

The introduction of a new quantized module requires the previously quantized modules to be updated simultaneously to achieve the optimal final quantized image. After the last module is quantized and introduced to the model, RPQ is transformed into end-to-end training eventually.  

\begin{table}[t]
\centering
\vspace{1.5mm}
\resizebox{1\columnwidth}{!}{
\begin{tabular}{c | c | c c}
\toprule
\rowcolor{iccvblue!30}
{Method} & {Bits} & {Params / M} ($\downarrow$ Ratio)  & {Ops / G} ($\downarrow$ Ratio)\\
\midrule
OSEDiff~\cite{wu2024one}      & W32A32 & 1,303 ($\downarrow$0\%) & 4,523 ($\downarrow$0\%) \\
\midrule
\rowcolor{iccvblue!10}
& W4A4 & 122 ($\downarrow$90.66\%) & 531 ($\downarrow$88.26\%) \\
\rowcolor{iccvblue!10}
& W3A3 & 92 ($\downarrow$92.94\%) & 398 ($\downarrow$91.19\%) \\
\rowcolor{iccvblue!10}
\multirow{-3}{*}{QArtSR}
& W2A2 & 62 ($\downarrow$95.21\%) & 266 ($\downarrow$94.12\%) \\
\bottomrule
\end{tabular}
}
\vspace{-3.5mm}
\caption{Params, Ops, and compression ratio of different quantization settings. Ops are computed with output size 512$\times$512.}
\label{tab:Compression ratio}
\vspace{-6.5mm}
\end{table}

RPQ's main disadvantage is that the later quantized modules receive fewer updates than earlier ones, resulting in insufficient training and potential performance degradation. Therefore, extended training (ET) is essential to ensure all modules are finetuned fully for better performance.  

\vspace{-1mm}
\subsection{Finetuning Quantizer Design of QArtSR}
\vspace{-1mm}
Inspired by recent quantization works~\cite{He2023EfficientDM,li2024svdqunat,smoothquant,OmniQuant}, we propose a specialized finetuning quantizer that minimizes the quantization error while adding minimal storage and computational overhead. This approach effectively balances the model performance and efficiency, making it highly suitable for resource-constrained environments.

For a linear layer with weight $W$, bias $B$, and activation $x$, we can design the finetuning quantizer as follows:
\begin{equation}
\begin{aligned}
    & y_q = Q_W(W)Q_A(x) + Q_B(B) \\
    &= L_1L_2x + Q_W(\phi(R+ F_1F_2))Q_A(\frac{x}{\phi}) + Q_B(B),
\end{aligned}
\label{equ: Finetune_quantizer}
\end{equation}
where $R = W - L_1L_2$. $Q_W$, $Q_B$, and $Q_A$ refer to the weight, bias, and activation quantizer. $L$ and $F$ are low-rank matrices. $\phi$ is the scale of equivalent transformation. 

In this quantizer, $\phi$, $L$, and $F$ are trainable parameters to improve the quantized layer's performance. $L$ serves as a low-cost full-precision skip connection between the input and quantized output. $F$ finetunes the residual matrix $R$, while $\phi$ can mitigate outliers and balance quantization pressure between the weights and activations.

During the finetuning process of RPQ, only quantization and finetuning parameters are updated, while the original model weights are frozen. The scale of trainable parameters is much smaller than that of model's weights, resulting in a relatively low finetuning cost of memory and time.

\begin{table*}[htbp]
\centering
\scriptsize
\resizebox{\textwidth}{!}
{
\begin{tabular}{c | c | c | c c c c c c c c}
\toprule
\rowcolor{iccvblue!30}
{Datasets} & {Bits} & {Methods} & PSNR$\uparrow$  & SSIM$\uparrow$ & LPIPS$\downarrow$ & DISTS$\downarrow$ & NIQE$\downarrow$ & MUSIQ$\uparrow$ & MANIQA$\uparrow$ & CLIP-IQA$\uparrow$ \\
\midrule
& \multirow{1}{*}{W32A32} 
&  OSEDiff~\cite{wu2024one} &25.27 & 0.7379 & 0.3027	& 0.1808 &4.355	& 67.43	&0.4766	&0.6835  \\
\cline{2-11}
& \multirow{7}{*}{W4A4} 
& MaxMin~\cite{jacob2017quantizationtrainingneuralnetworks} &16.01	&0.5033	&0.7016	&0.3668	&5.520	&31.26	&0.2771	&0.3099 \\
& & LSQ~\cite{Bhalgat2020LSQ+} &21.16	&0.6703	&0.4895	&0.2964	&8.906	&42.43	&0.2332	&0.2526 \\
& & Q-DM~\cite{li2024q} & 19.27	&0.5711	&0.4678	&0.3268	&\textcolor{blue}{5.387	}&\textcolor{blue}{48.04}	&\textcolor{blue}{0.2560}	&\textcolor{blue}{0.4706} \\
& & EfficientDM~\cite{He2023EfficientDM} &10.70	&0.5027	&0.8609	&0.3974	&5.846	&30.21	&0.2016	&0.2141 \\
& & PassionSR~\cite{zhu2024passionsr} &22.52	&0.6255	&0.4913	&0.3185	&5.706	&43.21	&0.2396	&0.3089 \\
& & SVDQuant~\cite{li2024svdqunat}  &\textcolor{red}{25.98}	&\textcolor{red}{0.7271}	&\textcolor{blue}{0.4618}	&\textcolor{blue}{0.2665}	&5.645	&42.84	&0.2555	&0.3986 \\
\rowcolor{iccvblue!10}
\cellcolor{white}& \cellcolor{white} & QArtSR~(ours) &\textcolor{blue}{24.50}	&\textcolor{blue}{0.7092}	&\textcolor{red}{0.3761}	&\textcolor{red}{0.2221}	&\textcolor{red}{4.808}	&\textcolor{red}{64.87}	&\textcolor{red}{0.4656}	&\textcolor{red}{0.6999} \\
\cline{2-11}
& \multirow{7}{*}{W2A2} 
& MaxMin~\cite{jacob2017quantizationtrainingneuralnetworks}  &14.88	&0.2181	&0.7263	&0.4584	&8.525	&33.12	&0.2187	&0.2190 \\
& & LSQ~\cite{Bhalgat2020LSQ+} &9.16	&0.0839	&0.8191	&0.5401	&15.48	&34.55	&\textcolor{blue}{0.3001}	&0.2997 \\
& & Q-DM~\cite{li2024q} &21.44	&\textcolor{blue}{0.6215}	&0.6229	&\textcolor{blue}{0.3063}	&9.081	&40.37	&0.2338	&0.2959 \\
& & EfficientDM~\cite{He2023EfficientDM} &11.81	&0.4496	&0.6879	&0.3684	&8.300	&35.38	&0.2453	&0.3464 \\
& & PassionSR~\cite{zhu2024passionsr} &14.56	&0.2330	&0.7609	&0.4904	&8.356	&37.54	&0.2376	&\textcolor{blue}{0.3694}\\
& & SVDQuant~\cite{li2024svdqunat}  &\textcolor{blue}{18.41}	&0.5941	&\textcolor{red}{0.4921} &0.3265	&\textcolor{blue}{5.953}	&\textcolor{red}{42.82}	&0.2447	&0.3659 \\
\rowcolor{iccvblue!10}
\cellcolor{white}\multirow{-15}{*}{RealSR~\cite{Ji_2020_CVPR_Workshops}} & \cellcolor{white} & QArtSR~(ours) &\textcolor{red}{24.92}	&\textcolor{red}{0.6353}	&\textcolor{blue}{0.5825}	&\textcolor{red}{0.2983}	&\textcolor{red}{5.942}	&\textcolor{blue}{40.98}	&\textcolor{red}{0.3102}	&\textcolor{red}{0.4865}\\
\midrule
& \multirow{1}{*}{W32A32} 
& OSEDiff~\cite{wu2024one} & 21.55	&0.6511	&0.2122	&0.1461	&4.622	&72.48	&0.4933	&0.6628 \\
\cline{2-11}
& \multirow{7}{*}{W4A4} 
& MaxMin~\cite{jacob2017quantizationtrainingneuralnetworks} &12.01	&0.1372	&0.7975	&0.4549	&5.674	&47.77	&0.2671	&0.2343 \\
& & LSQ~\cite{Bhalgat2020LSQ+} &18.19	&0.5616	&0.5172	&0.3366	&8.490	&50.80	&0.2793	&0.2747 \\
& & Q-DM~\cite{li2024q} &17.64	&0.5207	&0.4975	&0.3336	&5.052	&51.50	&0.2932	&0.4213 \\ 
& & EfficientDM~\cite{He2023EfficientDM} &10.90	&0.4528	&0.7672	&0.3958	&5.400	&38.17	&0.2303	&0.2327 \\
& & PassionSR~\cite{zhu2024passionsr} &19.20	&0.5325	&0.5680	&0.3440	&6.131	&46.99	&0.2617	&0.3083 \\
& & SVDQuant~\cite{li2024svdqunat}  &\textcolor{blue}{22.30}	&\textcolor{blue}{0.6113}	&\textcolor{blue}{0.3764}	&\textcolor{blue}{0.2493}	&\textcolor{blue}{4.708}	&\textcolor{blue}{55.98}	&\textcolor{blue}{0.3415}	&\textcolor{blue}{0.4339}\\
\rowcolor{iccvblue!10}
\cellcolor{white}& \cellcolor{white} & QArtSR~(ours) &\textcolor{red}{22.18}	&\textcolor{red}{0.6377}	&\textcolor{red}{0.2507}	&\textcolor{red}{0.1712}	&\textcolor{red}{4.227}	&\textcolor{red}{69.82}	&\textcolor{red}{0.4514}	&\textcolor{red}{0.6451}\\
\cline{2-11}
& \multirow{7}{*}{W2A2} 
& MaxMin~\cite{jacob2017quantizationtrainingneuralnetworks} &12.76	&0.3171	&0.8899	&0.4525	&9.417	&33.75	&0.2987	&0.2674 \\
& & LSQ~\cite{Bhalgat2020LSQ+} &8.39	&0.0797	&0.8500	&0.5337	&15.491	&44.37	&0.3031	&0.2918 \\
& & Q-DM~\cite{li2024q} &\textcolor{blue}{18.34}	&0.5802	&0.5559	&0.3448	&8.634	&48.88	&0.2791	&0.3157\\
& & EfficientDM~\cite{He2023EfficientDM} &11.69	&0.4116	&0.7075	&0.3910	&8.348	&42.06	&0.2672	&0.3479 \\  
& & PassionSR~\cite{zhu2024passionsr} &12.35	&0.1727	&0.8462	&0.4953	&9.636	&47.15	&\textcolor{blue}{0.3272}	&0.3491 \\
& & SVDQuant~\cite{li2024svdqunat}  &17.40	&\textcolor{red}{0.5687}	&\textcolor{blue}{0.4767}	&\textcolor{blue}{0.3102}	&\textcolor{blue}{6.063}	&\textcolor{blue}{50.06}	&0.3001	&\textcolor{blue}{0.3981}\\
\rowcolor{iccvblue!10}
\cellcolor{white}\cellcolor{white}\multirow{-15}{*}{Urban100~\cite{Huang-CVPR-2015}} & \cellcolor{white} & QArtSR~(ours) &\textcolor{red}{21.96}	&\textcolor{blue}{0.5682}	&\textcolor{red}{0.4299}	&\textcolor{red}{0.2720}	&\textcolor{red}{4.508}	&\textcolor{red}{55.48}	&\textcolor{red}{0.3878}	&\textcolor{red}{0.5229}\\
\midrule
& \multirow{1}{*}{W32A32} 
& OSEDiff~\cite{wu2024one}
& 24.95 & 0.7154	& 0.2325	& 0.1197	& 3.616 &	68.92	& 0.4340 &	0.6842 \\
\cline{2-11}
& \multirow{7}{*}{W4A4} 
& MaxMin~\cite{jacob2017quantizationtrainingneuralnetworks} &15.30	&0.5015	&0.7940	&0.4347	&5.260	&46.27	&0.2679	&0.2785 \\
& & LSQ~\cite{Bhalgat2020LSQ+} & 19.90	&0.6551	&0.5536	&0.3123	&8.011	&45.25	&0.2372	&0.3361\\
& & Q-DM~\cite{li2024q}  &19.40	&0.5873	&0.5003	&0.2748	&4.556	&49.20	&0.2511	&0.3857\\
& & EfficientDM~\cite{He2023EfficientDM} &10.82	&0.5031	&0.7858	&0.3914	&5.273	&32.41	&0.1989	&0.2325\\
& & PassionSR~\cite{zhu2024passionsr} &21.59	&0.6199	&0.5742	&0.3025	&5.865	&42.39	&0.2317	&0.3070 \\
& & SVDQuant~\cite{li2024svdqunat}  &\textcolor{red}{26.45}	&\textcolor{red}{0.7147}	&\textcolor{blue}{0.3565}	&\textcolor{blue}{0.2142}	&\textcolor{blue}{4.336}	&\textcolor{blue}{51.14}	&\textcolor{blue}{0.2980}	&\textcolor{blue}{0.4829} \\
\rowcolor{iccvblue!10}
\cellcolor{white}& \cellcolor{white} & QArtSR~(ours) &\textcolor{blue}{24.93}	&\textcolor{blue}{0.6890}	&\textcolor{red}{0.2914}	&\textcolor{red}{0.1638}	&\textcolor{red}{3.785}	&\textcolor{red}{67.98}	&\textcolor{red}{0.4395}	&\textcolor{red}{0.7220}\\
\cline{2-11}
& \multirow{7}{*}{W2A2} 
& MaxMin~\cite{jacob2017quantizationtrainingneuralnetworks} &14.50	&0.2320	&0.8159	&0.3660	&8.609	&33.22	&0.2374	&0.2781 \\
& & LSQ~\cite{Bhalgat2020LSQ+} &9.11	&0.1083	&0.8709	&0.5272	&13.73	&44.51	&0.3074	&0.3614 \\
& & Q-DM~\cite{li2024q} &20.03	&0.5681	&0.5858	&0.3274	&8.105	&42.63	&0.2342	&0.3727 \\
& & EfficientDM~\cite{He2023EfficientDM} &12.10	&0.4674	&0.7471	&0.3599	&7.936	&35.80	&0.2525	&0.3799\\
& & PassionSR~\cite{zhu2024passionsr} &13.57	&0.2195	&0.8486	&0.4940	&8.526	&\textcolor{blue}{46.53}	&\textcolor{blue}{0.3325}	&\textcolor{blue}{0.4326} \\
& & SVDQuant~\cite{li2024svdqunat} &\textcolor{blue}{18.99}	&\textcolor{blue}{0.6348}	&\textcolor{blue}{0.4947}	&\textcolor{blue}{0.2716}	&\textcolor{blue}{5.623}	&43.83	&0.2409	&0.3575 \\
\rowcolor{iccvblue!10}
\cellcolor{white}\cellcolor{white}\multirow{-15}{*}{DIV2K\_val~\cite{Agustsson_2017_CVPR_Workshops}}& \cellcolor{white} & QArtSR~(ours) &\textcolor{red}{25.51}	&\textcolor{red}{0.6479}	&\textcolor{red}{0.4124}	&\textcolor{red}{0.2386}	&\textcolor{red}{4.443}	&\textcolor{red}{50.04}	&\textcolor{red}{0.3439}	&\textcolor{red}{0.5589} \\
\cline{1-11}
\bottomrule
\end{tabular}
}
\vspace{-3.5mm}
\caption{Quantitative quantization experiments ($\times$4) results. The full-precision backbone is OSEDiff~\cite{wu2024one} ($T$=1,000). The best and second best results in the same setting are colored with \textcolor{red}{red} and \textcolor{blue}{blue}. W$w$A$a$ denotes $w$ bits weight and $a$ bits activation quantization.}
\vspace{-6.5mm}
\label{tab:Whole model quantization experiment results.}
\end{table*}

\section{Experiments}
\vspace{-2mm}
\subsection{Experiment Setup}
\vspace{-2mm}
\noindent \textbf{Data Construction.} We randomly extract 512 pairs of low-resolution (LR) and high-resolution (HR) images, each with a dimension of 128$\times$128, from the DIV2K\_train dataset~\cite{Agustsson_2017_CVPR_Workshops} to build the calibration set. For testing, we use datasets including RealSR~\cite{Ji_2020_CVPR_Workshops}, Urban100~\cite{Huang-CVPR-2015}, and DIV2K\_val~\cite{Agustsson_2017_CVPR_Workshops}.

\noindent \textbf{Evaluation Metrics.} We adopt reference-based evaluation metrics such as PSNR, SSIM~\cite{wang2004image}, LPIPS~\cite{zhang2018unreasonable}, and DISTS~\cite{ding2020image}. Additionally, we apply non-reference metrics, like NIQE~\cite{zhang2015feature}, MUSIQ~\cite{ke2021musiq}, MANIQA~\cite{yang2022maniqa}, and CLIPIQA~\cite{wang2023exploring}. All methods are evaluated using full-size images.

\noindent \textbf{Implementation Details.} Learning rate for QArtSR is set to $1$$\times$${10}^{-5}$. Experiments are conducted on RTX A6000, consuming 36.8 GB GPU memory and 6.8 hours GPU time, which are close to advanced PTQ methods~\cite{He2023EfficientDM, li2024svdqunat}.

\noindent \textbf{Compared Methods.} We compare QArtSR with several representative quantization methods: MaxMin~\cite{jacob2017quantizationtrainingneuralnetworks}, LSQ~\cite{esser2019learned}, Q-DM~\cite{li2024q}, EfficientDM~\cite{He2023EfficientDM}, PassionSR~\cite{zhu2024passionsr}, and SVDQuant~\cite{li2024svdqunat}. We adopt these quantization methods on OSEDiff~\cite{wu2024one} based on their released code. Q-DM, EfficientDM, and SVDQuant are newly proposed quantization methods for multi-step diffusion models. 
\vspace{-1mm}

\noindent \textbf{Compression Ratio.} We calculate the total model size (Params / M) and the number of operations (Ops / G), following the methodology adopted in prior quantization studies~\cite{quantsr}. The results, summarized in Tab.~\ref{tab:Compression ratio}, present the compression and acceleration ratios for the various settings with scaling factor 4. In 4-bit configuration, QArtSR achieves a compression ratio of 90.66\% and an acceleration ratio of 88.26\%, while those in 2-bit setting are 95.21\% and 94.12\%, compared to the FP OSEDiff model.

\begin{figure*}[!t]
\scriptsize
\centering

\begin{tabular}{cccccccc}
\hspace{-0.44cm}
\begin{adjustbox}{valign=t}
\begin{tabular}{c}
\includegraphics[width=0.185\textwidth]{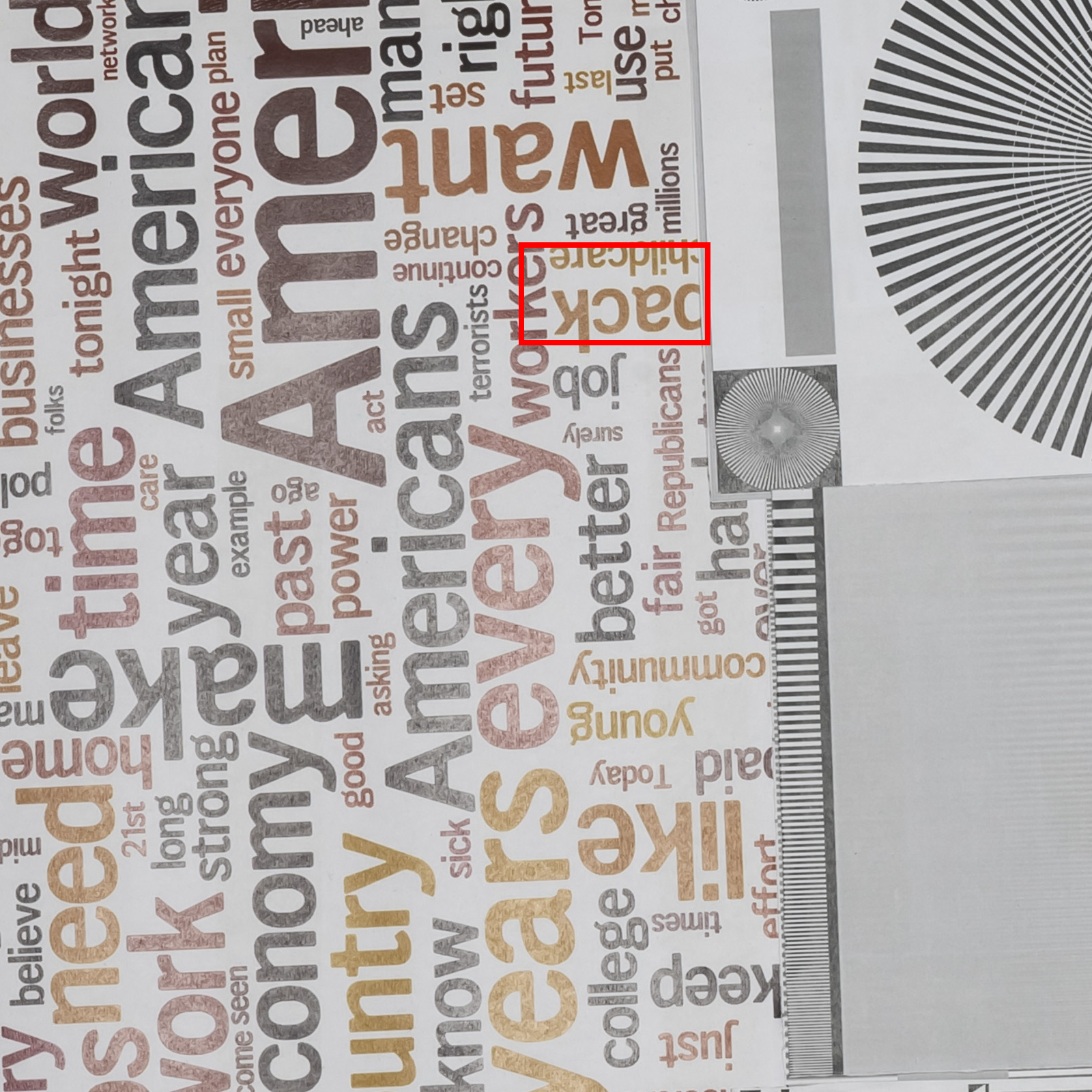}
\\
RealSR~\cite{Ji_2020_CVPR_Workshops}: Nikon\_050
\end{tabular}
\end{adjustbox}
\hspace{-0.46cm}
\begin{adjustbox}{valign=t}
\begin{tabular}{cccccc}
\includegraphics[width=0.154\textwidth]{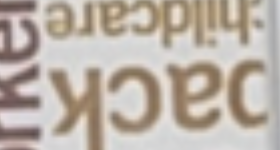} \hspace{-3.5mm} &
\includegraphics[width=0.154\textwidth]{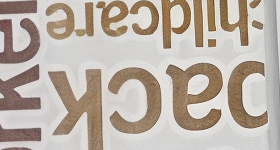} \hspace{-3.5mm} &
\includegraphics[width=0.154\textwidth]{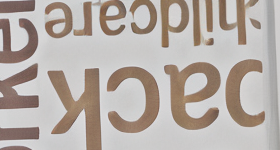} \hspace{-3.5mm} &
\includegraphics[width=0.154\textwidth]{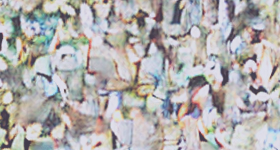} \hspace{-3.5mm} &
\includegraphics[width=0.154\textwidth]{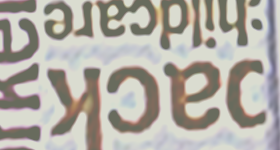} 
\hspace{-3.5mm} &
\\ 
LR / \#Steps / Bits \hspace{-3.5mm} &
DiffBIR~\cite{diffbir} / 50 / 32-bit\hspace{-3.5mm} &
OSEDiff~\cite{wu2024one} / 1 / 32-bit \hspace{-3.5mm} &
MaxMin~\cite{jacob2017quantizationtrainingneuralnetworks} / 1 / 4-bit \hspace{-3.5mm} &
LSQ~\cite{esser2019learned} / 1 / 4-bit \hspace{-3.5mm} &
\\
\includegraphics[width=0.154\textwidth]{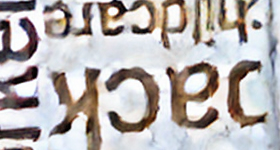} \hspace{-3.5mm} &
\includegraphics[width=0.154\textwidth]{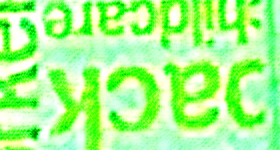} \hspace{-3.5mm} &
\includegraphics[width=0.154\textwidth]{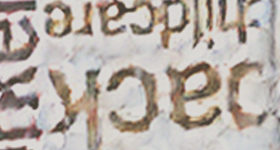} \hspace{-3.5mm} &
\includegraphics[width=0.154\textwidth]{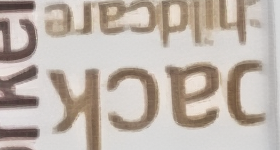} \hspace{-3.5mm} &
\includegraphics[width=0.154\textwidth]{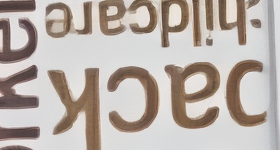} \hspace{-3.5mm} &
\\ 
Q-DM~\cite{li2024q}  / 1 / 4-bit \hspace{-3.5mm} &
EfficientDM~\cite{He2023EfficientDM}  / 1 / 4-bit \hspace{-3.5mm} &
PassionSR~\cite{zhu2024passionsr}  / 1 / 4-bit \hspace{-3.5mm} &
SVDQuant~\cite{li2024svdqunat}  / 1 / 4-bit \hspace{-3.5mm} &
QArtSR (ours)  / 1 / 4-bit \hspace{-3.5mm}
\\
\end{tabular}
\end{adjustbox}
\end{tabular}
\vspace{-0.5mm}

\begin{tabular}{cccccccc}
\hspace{-0.44cm}
\begin{adjustbox}{valign=t}
\begin{tabular}{c}
\includegraphics[width=0.185\textwidth]{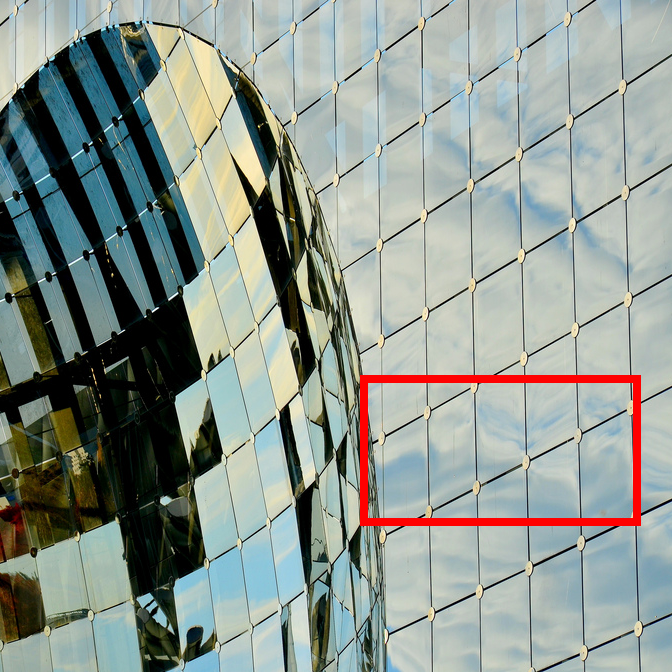}
\\
Urban100~\cite{Huang-CVPR-2015}: img029
\end{tabular}
\end{adjustbox}
\hspace{-0.46cm}
\begin{adjustbox}{valign=t}
\begin{tabular}{cccccc}
\includegraphics[width=0.154\textwidth]{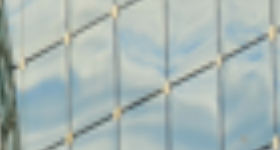} \hspace{-3.5mm} &
\includegraphics[width=0.154\textwidth]{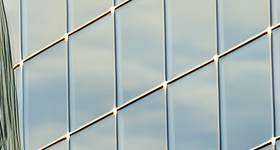} \hspace{-3.5mm} &
\includegraphics[width=0.154\textwidth]{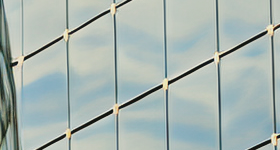} \hspace{-3.5mm} &
\includegraphics[width=0.154\textwidth]{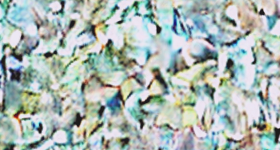} \hspace{-3.5mm} &
\includegraphics[width=0.154\textwidth]{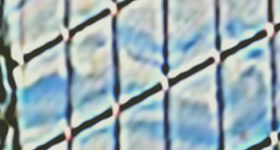} \hspace{-3.5mm} &
\\ 
LR / \#Steps / Bits \hspace{-3.5mm} &
DiffBIR~\cite{diffbir} / 50 / 32-bit\hspace{-3.5mm} &
OSEDiff~\cite{wu2024one} / 1 / 32-bit \hspace{-3.5mm} &
MaxMin~\cite{jacob2017quantizationtrainingneuralnetworks} / 1 / 4-bit \hspace{-3.5mm} &
LSQ~\cite{esser2019learned} / 1 / 4-bit \hspace{-3.5mm} &
\\
\includegraphics[width=0.154\textwidth]{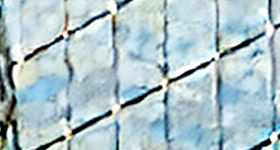} \hspace{-3.5mm} &
\includegraphics[width=0.154\textwidth]{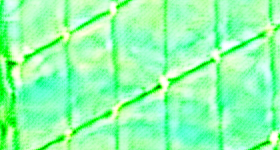} \hspace{-3.5mm} &
\includegraphics[width=0.154\textwidth]{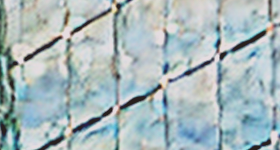} \hspace{-3.5mm} &
\includegraphics[width=0.154\textwidth]{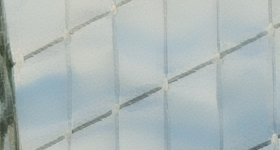} \hspace{-3.5mm} &
\includegraphics[width=0.154\textwidth]{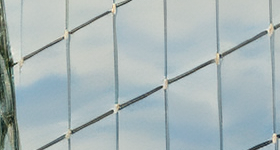} \hspace{-3.5mm} &
\\ 
Q-DM~\cite{li2024q} / 1 / 4-bit \hspace{-3.5mm} &
EfficientDM~\cite{He2023EfficientDM}  / 1 / 4-bit \hspace{-3.5mm} &
PassionSR~\cite{zhu2024passionsr}  / 1 / 4-bit \hspace{-3.5mm} &
SVDQuant~\cite{li2024svdqunat}  / 1 / 4-bit \hspace{-3.5mm} &
QArtSR (ours)  / 1 / 4-bit \hspace{-3.5mm}
\\
\end{tabular}
\end{adjustbox}
\end{tabular}
\vspace{-0.5mm}

\begin{tabular}{cccccccc}
\hspace{-0.44cm}
\begin{adjustbox}{valign=t}
\begin{tabular}{c}
\includegraphics[width=0.185\textwidth]{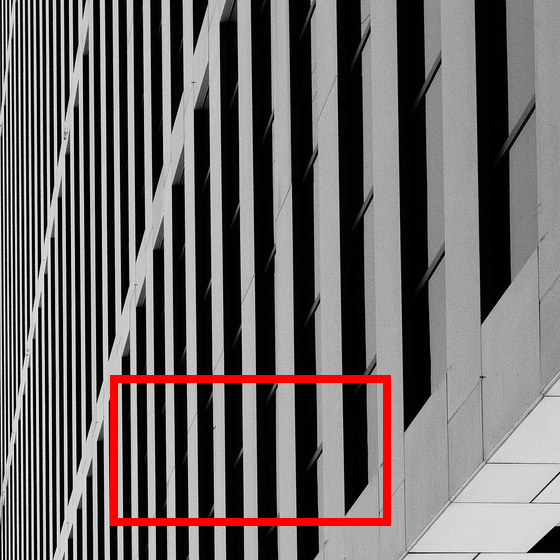}
\\
Urban100~\cite{Huang-CVPR-2015}: img011
\end{tabular}
\end{adjustbox}
\hspace{-0.46cm}
\begin{adjustbox}{valign=t}
\begin{tabular}{cccccc}
\includegraphics[width=0.154\textwidth]{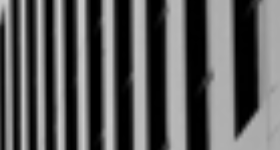} \hspace{-3.5mm} &
\includegraphics[width=0.154\textwidth]{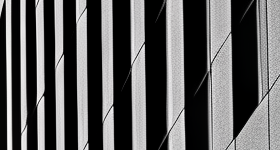} \hspace{-3.5mm} &
\includegraphics[width=0.154\textwidth]{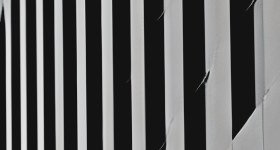} \hspace{-3.5mm} &
\includegraphics[width=0.154\textwidth]{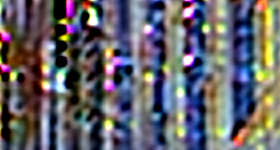} \hspace{-3.5mm} &
\includegraphics[width=0.154\textwidth]{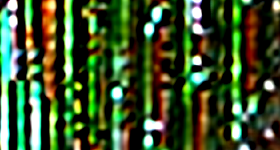} \hspace{-3.5mm} &
\\ 
LR / \#Steps / Bits \hspace{-3.5mm} &
DiffBIR~\cite{diffbir} / 50 / 32-bit\hspace{-3.5mm} &
OSEDiff~\cite{wu2024one} / 1 / 32-bit \hspace{-3.5mm} &
MaxMin~\cite{jacob2017quantizationtrainingneuralnetworks} / 1 / 2-bit \hspace{-3.5mm} &
LSQ~\cite{esser2019learned} / 1 / 2-bit \hspace{-3.5mm} &
\\
\includegraphics[width=0.154\textwidth]{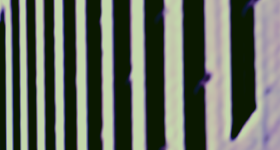} \hspace{-3.5mm} &
\includegraphics[width=0.154\textwidth]{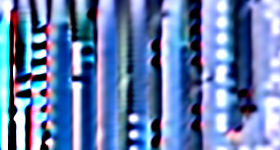} \hspace{-3.5mm} &
\includegraphics[width=0.154\textwidth]{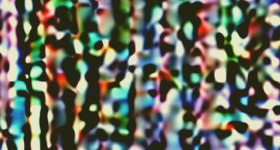} \hspace{-3.5mm} &
\includegraphics[width=0.154\textwidth]{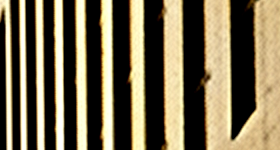} \hspace{-3.5mm} &
\includegraphics[width=0.154\textwidth]{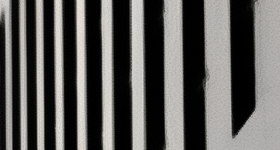} \hspace{-3.5mm} &
\\ 
Q-DM~\cite{li2024q} / 1 / 2-bit \hspace{-3.5mm} &
EfficientDM~\cite{He2023EfficientDM}  / 1 / 2-bit \hspace{-3.5mm} &
PassionSR~\cite{zhu2024passionsr}  / 1 / 2-bit \hspace{-3.5mm} &
SVDQuant~\cite{li2024svdqunat}  / 1 / 2-bit \hspace{-3.5mm} &
QArtSR (ours)  / 1 / 2-bit \hspace{-3.5mm}
\\
\end{tabular}
\end{adjustbox}
\end{tabular}
\vspace{-0.5mm}

\begin{tabular}{cccccccc}
\hspace{-0.44cm}
\begin{adjustbox}{valign=t}
\begin{tabular}{c}
\includegraphics[width=0.185\textwidth]{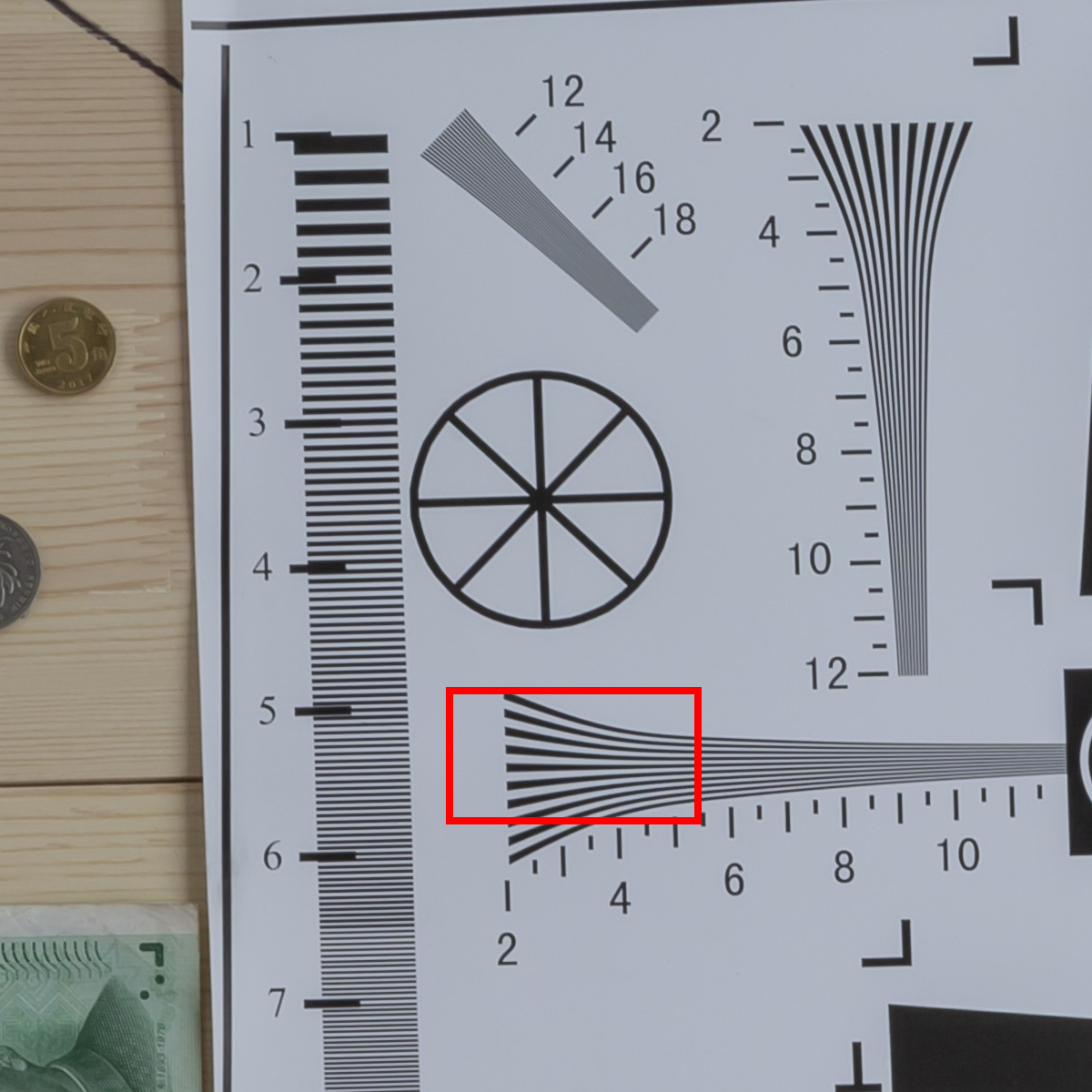}
\\
RealSR~\cite{Ji_2020_CVPR_Workshops}: Canon\_045
\end{tabular}
\end{adjustbox}
\hspace{-0.46cm}
\begin{adjustbox}{valign=t}
\begin{tabular}{cccccc}
\includegraphics[width=0.154\textwidth]{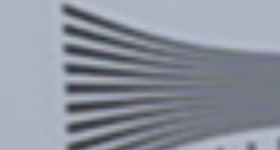} \hspace{-3.5mm} &
\includegraphics[width=0.154\textwidth]{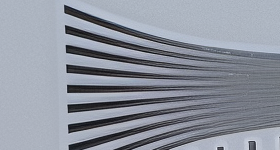} \hspace{-3.5mm} &
\includegraphics[width=0.154\textwidth]{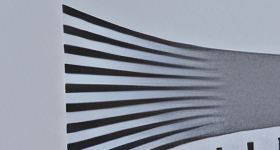} \hspace{-3.5mm} &
\includegraphics[width=0.154\textwidth]{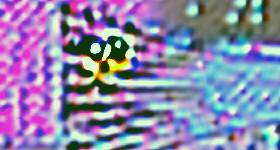} \hspace{-3.5mm} &
\includegraphics[width=0.154\textwidth]{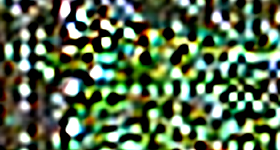} 
\hspace{-3.5mm} &
\\ 
LR / \#Steps / Bits \hspace{-3.5mm} &
DiffBIR~\cite{diffbir} / 50 / 32-bit\hspace{-3.5mm} &
OSEDiff~\cite{wu2024one} / 1 / 32-bit \hspace{-3.5mm} &
MaxMin~\cite{jacob2017quantizationtrainingneuralnetworks} / 1 / 2-bit \hspace{-3.5mm} &
LSQ~\cite{esser2019learned} / 1 / 2-bit \hspace{-3.5mm} &
\\
\includegraphics[width=0.154\textwidth]{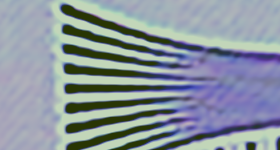} \hspace{-3.5mm} &
\includegraphics[width=0.154\textwidth]{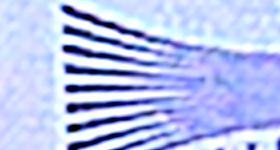} \hspace{-3.5mm} &
\includegraphics[width=0.154\textwidth]{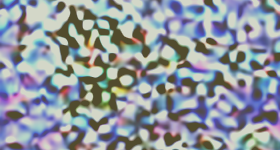} 
\hspace{-3.5mm} &
\includegraphics[width=0.154\textwidth]{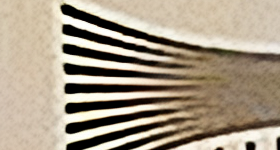} 
\hspace{-3.5mm} &
\includegraphics[width=0.154\textwidth]{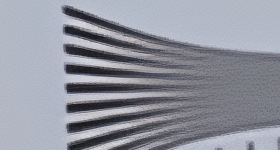} \hspace{-3.5mm} &
\\ 
Q-DM~\cite{li2024q} / 1 / 2-bit \hspace{-3.5mm} &
EfficientDM~\cite{He2023EfficientDM}  / 1 / 2-bit \hspace{-3.5mm} &
PassionSR~\cite{zhu2024passionsr}  / 1 / 2-bit \hspace{-3.5mm} &
SVDQuant~\cite{li2024svdqunat}  / 1 / 2-bit \hspace{-3.5mm} &
QArtSR (ours)  / 1 / 2-bit \hspace{-3.5mm}
\\
\end{tabular}
\end{adjustbox}
\end{tabular}
\vspace{-3.5mm}
\caption{Visual comparison ($\times 4$) of high-resolution images, full-precision model outputs, and various quantization methods in challenging cases under W4A4 and W2A2 settings. QArtSR demonstrates notable visual superiority over other approaches.}
\vspace{-3mm}
\label{fig:visual-whole-model}
\end{figure*}

\begin{table*}[t]
\centering
\scriptsize
\resizebox{\textwidth}{!}{
\begin{tabular}{c | c c c | c c c c c c c c c}
\toprule
\rowcolor{iccvblue!30}
&  &\multicolumn{2}{c|}{RPQ} & \multicolumn{8}{c}{RealSR~\cite{Ji_2020_CVPR_Workshops}} \\
\rowcolor{iccvblue!30}
\multirow{-2}{*}{Bits} & \multirow{-2}{*}{TRQ} & $\text{RPQ}^*$ & ET & PSNR$\uparrow$  & SSIM$\uparrow$ & LPIPS$\downarrow$ & DISTS$\downarrow$ & NIQE$\downarrow$ & MUSIQ$\uparrow$ & MANIQA$\uparrow$ & CLIP-IQA$\uparrow$ \\
\midrule
&   & & &16.01	&0.5033	&0.7016	&0.3668	&5.520	&31.26	&0.2771	&0.3099\\
&\checkmark   & & &\textcolor{red}{26.31}	&\textcolor{red}{0.7328}	&0.4423	&0.3035	&7.280	&38.49	&0.2010	&0.3941 \\
& & \checkmark & &22.36	&0.6735	&0.5248	&0.3489	&5.879	&47.57	&0.2962	&0.5601 \\
& \checkmark & \checkmark & &\textcolor{blue}{25.17}	&\textcolor{blue}{0.7171}	&\textcolor{blue}{0.4035}	&0.2498	&\textcolor{blue}{5.105}	&54.93	&0.3584	&0.6184  \\
&\checkmark  & & \checkmark &24.79	&0.7083	&0.4113	&\textcolor{blue}{0.2447}	&5.130	&\textcolor{blue}{59.83}	&\textcolor{blue}{0.4303}	&\textcolor{blue}{0.6516}  \\
\rowcolor{iccvblue!10}
\cellcolor{white}\multirow{-6}{*}{W4A4} &\checkmark   & \checkmark & \checkmark &24.50	&0.7092	&\textcolor{red}{0.3761}	&\textcolor{red}{0.2221} &\textcolor{red}{4.808}	&\textcolor{red}{64.87} &\textcolor{red}{0.4656}	&\textcolor{red}{0.6999} \\
\cline{1-12}
\bottomrule
\end{tabular}
}
\vspace{-3.mm}
\caption{Ablation study ($\times 4$) on the key components of our proposed method: TRQ and RPQ. RPQ consists of two stages: per-module quantization training ($\text{RPQ}^*$) and extended training (ET). The best and second-best results in this setting are highlighted in \textcolor{red}{red} and \textcolor{blue}{blue}. }
\vspace{-5.5mm}
\label{tab:ablation experiments}
\end{table*}

\subsection{Main Results}
\vspace{-3.5mm}
\noindent \textbf{Quantitative Results.} Table~\ref{tab:Whole model quantization experiment results.} presents the quantitative results of 4-bit and 2-bit settings. QArtSR significantly outperforms other quantization methods in both settings. In 4-bit setting, QArtSR performs comparably with the FP model and even surpasses it in certain metrics. While QArtSR is outperformed by SVDQuant~\cite{li2024svdqunat} in some metrics, such as PSNR and SSIM, QArtSR maintains a significant advantage across a broader range of evaluation criteria. Other 4-bit models show a large performance gap compared to the FP model, highlighting QArtSR’s effectiveness. In 2-bit setting, other models suffer severe performance drops, while QArtSR achieves the smallest performance gap and proves its robustness in ultra-low-bit quantization.

\begin{figure*}[t]
\centering
\includegraphics[width=\textwidth]{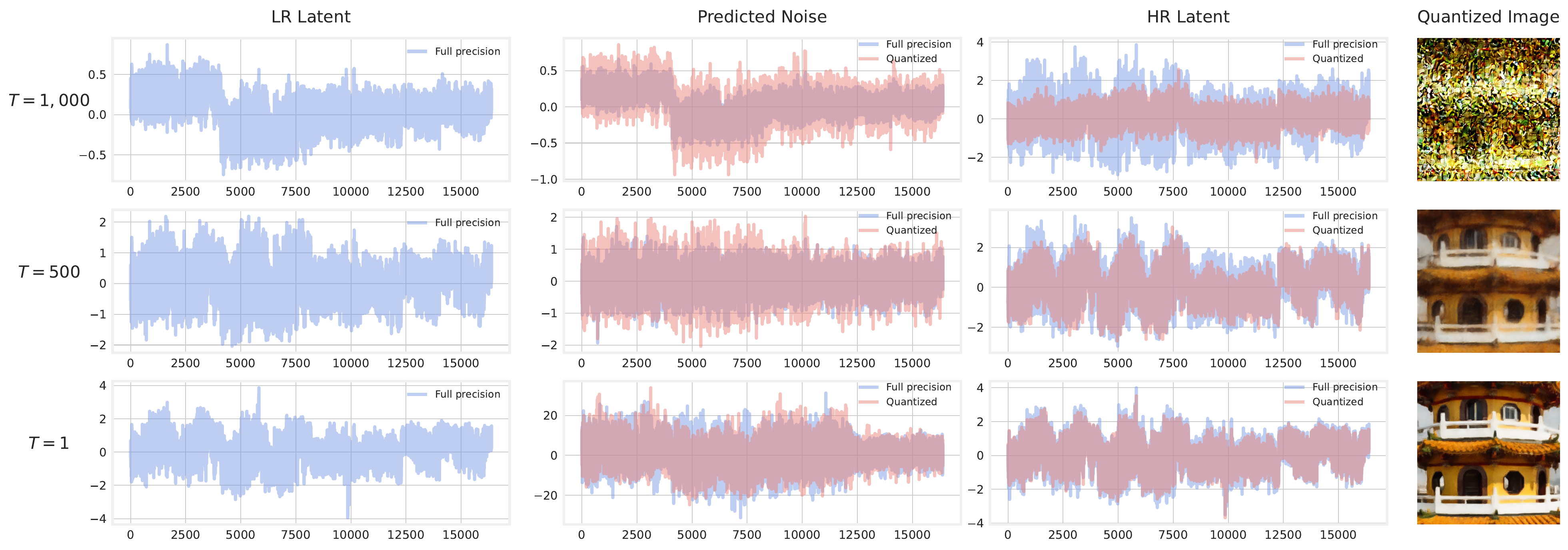}
\vspace{-8mm}
\caption{Visualization of three essential staged outputs and the quantized image in OSEDiff~\cite{wu2024one} under $T$=1,000,\ 500,\ 1 respectively.}
\label{fig:data_visual}
\vspace{-5.5mm}
\end{figure*}

\noindent\textbf{Visual Comparison.} Figure~\ref{fig:visual-whole-model} shows a visual comparison ($\times 4$), highlighting many challenging cases for better and clearer contrast. QArtSR produces sharper details and more refined textures than other comparison methods. There is a minor gap between the 4-bit QArtSR and the full-precision model, indicating QArtSR's outstanding performance. Compared with 4-bit quantized models, the performance distinction among the 2-bit quantization methods is more pronounced. Although the 2-bit QArtSR exhibits obvious performance degradation, it outperforms other comparison methods by a large margin. It still retains much expression ability for clearer HR images while the other 2-bit quantized models almost fail to generate HR images.

\vspace{-1.2mm}
\subsection{Feature Visualization}
\vspace{-1.2mm}
In Fig.~\ref{fig:data_visual}, we present three key staged features along with the corresponding quantized images for $T$=1,000,\ 500,\ 1, respectively. Aiming at illustrating the effects of model quantization in Eq.~\eqref{equ: OSDSR-LH}, we utilize the output from the full-precision VAE encoder as LR latent, providing a clear reference for comparison between FP and quantized models. It is evident that the quantization error of the predicted noise is the largest when $T$=1 while the quantization errors of the HR latent and the quantized image are the smallest instead. This consistency between experimental results and our theory further reinforces the validity of our approach.

\vspace{-1.2mm}
\subsection{Ablation Study}
\vspace{-1.2mm}
\noindent\textbf{Timestep Retraining for Quantization (TRQ).} The results presented in Tab.~\ref{tab:Step Study} indicate that the quantized model under timestep $T$ = 1 achieves superior performance, especially at ultra-low-bit settings. It highlights the effectiveness of retraining the backbone model with better timestep for quantization. The improvements are especially noticeable for aggressive quantization, confirming that the choice of timestep plays a crucial role in preserving model performance. Ablation study results in Tab.~\ref{tab:ablation experiments} also demonstrate that the introduction of TRQ brings great enhancement to the baseline and $\text{RPQ}^*$ on most of the evaluation metrics.

\noindent\textbf{Reversed Per-module Quantization ($\text{RPQ}^*$).} To evaluate the effects of $\text{RPQ}^*$, we conduct two comparative experiments, as shown in Tab.~\ref{tab:ablation experiments}. $\text{RPQ}^*$ significantly outperforms the baseline, showing substantial improvements across most of the evaluation metrics. Furthermore, when $\text{RPQ}^*$ is integrated into the TRQ framework, it contributes to notable gains in performance on nearly all metrics, highlighting its effectiveness in enhancing the overall model performance.

\noindent\textbf{Extended Training (ET).} As shown in Tab.~\ref{tab:ablation experiments}, comparing TRQ and TRQ+ET reveals that the incorporation of extended training significantly enhances the performance of the quantized model. Furthermore, by combining ET with TRQ+$\text{RPQ}^*$, which involves extending the training epochs after $\text{RPQ}^*$, we observe a substantial improvement in the model's performance. The extended training enables the quantized model to recover more effectively from the performance degradation caused by quantization, leading to a more robust and efficient quantized model.

\vspace{-2mm}
\section{Conclusion}
\vspace{-2mm}
We propose QArtSR, a novel quantization method for one-step diffusion-based image SR models. First, we investigate impact of timestep values on OSDSR quantization. We propose the timestep retraining quantization (TRQ) strategy, easing subsequent quantization and enabling high performance even under ultra-low-bit settings. Additionally, we propose a reversed per-module quantization (RPQ) technique relative to the inference sequence, allowing joint optimization of module and image loss. We then apply extended training after per-module stage to ensure that quantized modules are fully finetuned. Quantization experiments demonstrate that QArtSR delivers perceptual quality compared with FP models at 4-bit and preserves most of its performance even at 2-bit. It surpasses recent diffusion quantization methods, establishing itself as a strong candidate for OSDSR quantization. This work lays the foundation for the future advancements in OSDSR quantization and the practical deployment of high-performance SR models.

{
    \small
    \bibliographystyle{ieeenat_fullname}
    \bibliography{main}
}


\end{document}